\pdfoutput=1
\documentclass[letterpaper, 10 pt, conference]{ieeeconf}  % Comment this line out if you need a4paper

\IEEEoverridecommandlockouts                              % This command is only needed if 
\newcommand{\yash}[1]{}
\newcommand{\red}[1]{}
\newcommand{\tj}[1]{}
\newcommand{\kar}[1]{}
\newcommand{\chris}[1]{}
\newcommand{\task}[1]{}

\newcommand{\qq}[1]{PIXER}

\usepackage{graphicx} % for pdf, bitmapped graphics files
\usepackage{multirow}
\usepackage{multicol}
\usepackage{array,booktabs}
\usepackage[skins,theorems]{tcolorbox}
\usepackage{hyperref}
\usepackage{tabularx}
\usepackage{xcolor}
\usepackage{caption}
\usepackage[font=small,labelfont=bf]{caption}
\usepackage{subcaption}
\usepackage[font=small,labelfont=bf]{subcaption}
\usepackage{float}
\usepackage{soul}
\title{\LARGE \bf
Learning Visual Information Utility with PIXER
}

\usepackage{graphicx}
\usepackage{etoolbox}
\newcommand{\insertfig}{
\includegraphics[trim=1.25cm 1.25cm 1.25cm 1.25cm, clip=true,width=\textwidth]{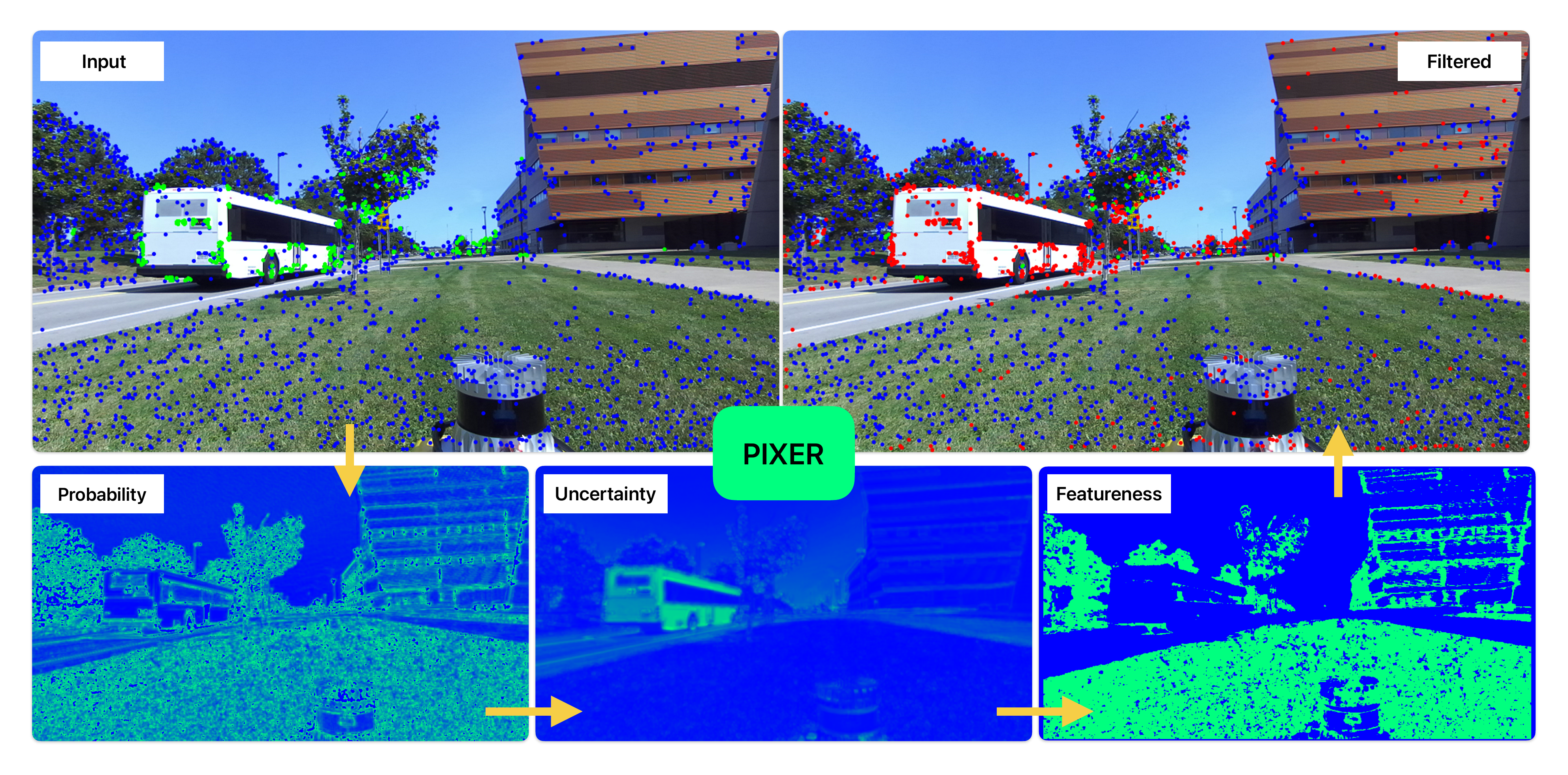}
\captionof{figure}{
In this work, we introduce the concept of ``Featureness'': the inherent interest and reliability of visual information for scene understanding in autonomous systems. Using a foundation of feature interest-point learning, \qq{} learns to output pixel-wise probability and uncertainty information that estimates a definition of featureness tailorable to a wide range of applications. \qq{} formulates a generalization on stochastic Bayesian learning to estimate featureness in a lightweight, single-shot architecture, where we demonstrate immediate utility by improving performance in visual odometry through the selective filtering of numerous front-end feature types. Examples of different feature types are shown in the top left image (SIFT blue, ORB green), while the top right shows these keypoints after \qq{} selection (red features are removed). The bottom displays probability ($P$) and uncertainty ($U$) heatmaps output by \qq{} with featureness mask ($F$) generated by selectively fusing $P$ and $U$.
}
\label{fig:intro}
}% define the image

\makeatletter
\apptocmd{\@maketitle}{\centering\insertfig}{}{}% insert the figure after authors
\makeatother
\author{Yash Turkar$^{*}$, Timothy Chase Jr$^{*}$, Christo Aluckal, Karthik Dantu% <-this % stops a space
\thanks{$^{*}$Authors contributed equally. \href{https://github.com/droneslab/PIXER}{{\color{blue}Project page}}. All authors are with the Department of Computer Science and Engineering, University at Buffalo, Buffalo, NY 14260, USA
        {\tt\scriptsize \{yashturk, tbchase, christoa, kdantu\}@buffalo.edu}}%
}

\begin{document}
\maketitle

\thispagestyle{empty}
\pagestyle{empty}

\begin{abstract}
Accurate feature detection is fundamental for various computer vision tasks including autonomous robotics, 3D reconstruction, medical imaging, and remote sensing. Despite advancements in enhancing the robustness of visual features, no existing method measures the utility of visual information before processing by specific feature-type algorithms. To address this gap, we introduce \qq{} and the concept of ``Featureness'', which reflects the inherent interest and reliability of visual information for robust recognition independent of any specific feature type. Leveraging a generalization on Bayesian learning, our approach quantifies both the probability and uncertainty of a pixel's contribution to robust visual utility in a single-shot process, avoiding costly operations such as Monte Carlo sampling, and permitting customizable featureness definitions adaptable to a wide range of applications. We evaluate \qq{} on visual odometry with featureness selectivity, achieving an average of 31\% improvement in RMSE trajectory with 49\% fewer features. Upon publication, \qq{}'s code, models, datasets, and detailed documentation will be made available.

% achieving comparable rotation and translation estimates with 49\% fewer features and a 31\% improvement in RMSE trajectory. Upon publication, \qq{}'s code, models, datasets, and detailed documentation will be made available.
% \yash{Broad application / not only feature reduction}}
% Accurate feature detection is fundamental for various tasks including autonomous robotics, 3D reconstruction, medical imaging, and remote sensing. Despite advancements in enhancing the robustness of visual features, no existing method measures the utility of visual information before processing by specific feature algorithms. To address this gap, we introduce \qq{} and the concept of "Featureness", which reflects the inherent interest and reliability of visual information for robust feature recognition. Leveraging a generalized Bayesian learning framework, our approach quantifies both the probability and uncertainty of a pixel's contribution to robust visual utility in a single-shot process, avoiding costly operations such as Monte Carlo sampling, and permitting customizable Featureness definitions adaptable to a wide range of applications. Towards perception safety, we evaluate \qq{} on visual odometry achieving comparable rotation and translation estimates with 51.5\% fewer features and a 33.8\% improvement in RMSE trajectory. Upon publication, \qq{}'s code, models, datasets, and detailed documentation will be available at \href{https://github.com/droneslab}{{\color{blue}github.com/droneslab}}.
\end{abstract}

% This paper introduces \qq{}, a learning-based framework for measuring "featureness" – the inherent interest and reliability of visual information for robust feature recognition, regardless of specific front-end feature extraction techniques. Utilizing a generalization of Bayesian learning, our approach quantifies both the probability and variance of a pixel's contribution to robust visual utility, all with a single-shot framework that avoids computationally expensive statistical inference operations such as Monte Carlo dropout. Our method is not limited to feature matching or estimation; we demonstrate its effectiveness in image space reduction by pixel-wise probability and uncertainty analysis. In a prototypical application such as visual odometry, this results in comparable rotation and translation estimation with significantly fewer features with a mean reduction of 51.5\% while achieving a RMSE trajectory improvement of 33.8\%, highlighting the broad applicability of our approach.
% Our implementation of \qq{}, custom real-world datasets, and detailed documentation on how to integrate the pipeline can be found at  \href{https://github.com/droneslab}{{\color{blue}github.com/droneslab}}
% upon publication. 
% \yash{Abstract needs to be rewritten}

%%%%%%%%%%%%%%%%%%%%%%%%%%%%%%%%%%%%%%%%%%%%%%%%%%%%%%%%%%%%%%%%%%%%%%%%%%%%%%%%
\section{Introduction}\label{intro}

Uncertainty plays a crucial role in autonomous systems by enabling more robust and accurate estimation. Bayesian estimation~\cite{barfoot_state_2017} techniques leverage sensor uncertainties, often represented as error covariances, to quantify the confidence in measurements. Sensors like IMUs typically provide these uncertainties directly, allowing for straightforward incorporation into Bayesian filters and optimization methods~\cite{dellaert_factor_2017}. However, estimating uncertainty from other sensors, particularly vision-based ones, is significantly more challenging.  Image data inherently contains complex ambiguities and noise patterns that are difficult to model precisely. Learning effective methods for quantifying image uncertainty, or uncertainty in the feature measurements, can significantly improve performance by providing a more cohesive understanding of the surrounding environment, its complexity, and its relation to the robot's state.
% more reliable information about the robot's pose and the surrounding environment.

Recent advances in deep learning have significantly improved image feature extraction and matching, crucial for visual applications such as Simultaneous Localization and Mapping (SLAM), 3D reconstruction, and visual place recognition (VPR). Neural networks can learn to reason about the complexities within images in a wide variety of environments and predict reliable feature points that are re-identifiable over time~\cite{detone_superpoint_2018,christiansen_unsuperpoint_2019,gleize_silk_2023,xiao_real-time_2024}. However, a common limitation across both traditional and learned feature extraction methods is the lack of uncertainty understanding as it relates to a given feature.  Most approaches fail to provide any measure of feature confidence, leaving us without a way to assess which features are more trustworthy and potentially mitigate errors in visual correspondence and estimation tasks. We argue that interest point probabilities inherent in feature learning are insufficient to adequately quantify the feature space, and an absence of uncertainty quantification represents an important gap in current research that highlights the need for methods that can characterize the quality of visual information for reliable and explainable robot perception. 

% take per-feature uncertainty into account for more reliable and explainable robot perception.

Bayesian Neural Networks (BNNs) are a recent machine learning advancement that offers uncertainty quantification alongside predictions. Unlike traditional neural networks which assign single values to weights, BNNs treat each weight as a probability distribution. During training, these distributions are updated based on the input data, allowing the network to learn not only the most likely output but also the range of possible outcomes and their associated probabilities.  However, obtaining accurate variance predictions from BNNs typically requires multiple inferences via Monte Carlo methods, which can be computationally expensive and impractical for real-time applications. 

This paper presents \qq{}, an approach to image understanding capable of simultaneously predicting both the probability and the uncertainty of pixel feature-likelihood in a lightweight, single-shot network. We formulate a generalization over Bayesian feature point learning to enable probabilistic pixel-wise predictions, incorporating a specialized \textit{uncertainty head} attached to a baseline detector that directly outputs uncertainty estimates without relying on computationally expensive Monte Carlo methods. 
% Building upon the SILK~\cite{gleize_silk_2023} network backbone, we replace its traditional layers with Bayesian counterparts, enabling probabilistic predictions.  
% Furthermore, we incorporate a specialized \textit{uncertainty head} that directly outputs variance without relying on computationally expensive Monte Carlo methods.
The fusion of feature probability and uncertainty predictions defines our concept of {\it featureness}, a general and versatile measure of how likely a pixel is to harbor a distinctive feature, paving the way for more robust and reliable visual perception for autonomous robots. This paper makes the following contributions:
\begin{itemize}
    \item We introduce a novel learning framework, \qq{} which predicts per-pixel visual utility of an input image. PIXER formulates a generalization on stochastic BNNs and outputs probability and uncertainty maps in a single-shot architecture.
    \item We define featurneess as the visual information utility, which is a pixel's inherent interest and reliability for robust perception.
    \item We demonstrate the utility of \qq{} for robot perception, improving RMSE in the visual odometry task by an average of 31\% with 49\% fewer features across three datasets and eight feature algorithms, with slight improvements in computational cost (0.63\%).
    
    % while using considerably less data (i.e., features) overall. 
    % and compute time in visual odometry pipelines 
    
    % We show with extensive experimentation that \qq{} visual odometry provides consistent improvement in RMSE and compute time while using considerably less data (i.e., features) overall.
\end{itemize}

% Removing the following para

% Feature extraction and matching from images have revolutionized numerous applications, including Simultaneous Localization and Mapping (SLAM), Visual Odometry (VO), and Structure from Motion (SfM). These techniques empower robots to accurately perceive and navigate their surroundings, a fundamental requirement for autonomy. Early approaches focused on extracting features based on intensity gradients and corners, but recent advancements have shifted towards learning-based methods for feature extraction. This shift leverages the power of deep learning to identify robust and discriminative features, crucial for applications like SLAM which demand reliable performance even in challenging environments.

% \task{Cite estimation / SLAM papers}

% \begin{figure}
%   \centering
%   \includegraphics[trim=1cm 1cm 1cm 1cm,clip=true,width=.9\columnwidth]{images/ICRA_Maps/Pallet_Combo_v6.pdf}
%   \caption{}
%   \label{intro_fig}
%   \vspace{-0.7cm}
% \end{figure}

\section{Related Work}\label{related}

\subsection{Uncertainty in Robot Estimation}
Probabilistic methods have transformed autonomous systems by incorporating uncertainty into state estimations. Techniques like filtering~\cite{bloesch_robust_2015,tanskanen_semi-direct_2015,geneva_openvins_2020,li_improving_2012} and factor graph optimization~\cite{qin_vins-mono_2018,shan_lio-sam_2020,shan_lego-loam_2018,campos_orb-slam3_2020,cioffi_tightly-coupled_2020} fuse uncertainty from robot motion models and sensor readings, proving effective in visual and visual-inertial systems like ORB-SLAM~\cite{campos_orb-slam3_2020}, VINS-Mono~\cite{qin_vins-mono_2018}, and OpenVINS~\cite{geneva_openvins_2020}. Monocular depth uncertainty over landmarks is used for active view planning~\cite{dai_uncertainty-driven_2021}, while learning-based approaches use dropout and LSTM layers to estimate uncertainty in temporal image pose predictions~\cite{costante_uncertainty_2020}. Bayesian convolutional networks apply Monte Carlo Dropout (MC-Dropout~\cite{gal_dropout_2016}) for per-pixel uncertainty in segmentation~\cite{ferianc_combinet_2021} tasks, and self-supervised training with homography transformation and decoder networks predict uncertainty for Micro Aerial Vehicles (MAVs)~\cite{xu_cuahn-vio_2022}. Conformal inference has been utilized to estimate visual odometry uncertainty using prediction intervals and MC-Dropout~\cite{stutts_lightweight_2023}, while factor graphs recover relative covariance estimates from implicit network layers~\cite{nir_designing_2024}.

\subsection{Feature Learning}
Generalized feature point learning has garnered much attention in recent years. SuperPoint~\cite{detone_toward_2017, detone_superpoint_2018} uses a convolutional neural network for detecting salient corners and trains self-supervised on synthetic data, later adapted to real-world imagery. Self-Improving Visual Odometry~\cite{detone_self-improving_2018} extends this idea by adding a ``stability'' network while UnSuperPoint~\cite{christiansen_unsuperpoint_2019} utilizes siamese networks for an unsupervised learning approach. GLAMPoints~\cite{truong_glampoints_2019} uses image transformations to create training pairs, while R2D2~\cite{revaud_r2d2_2019} incorporates reliability and repeatability heads to treat feature matching as a metric learning problem. Other architectures and learning paradigms have been proposed, including probabilistic reinforcement training (DISK)~\cite{tyszkiewicz_disk_2020}, transformers (LoFTR)~\cite{sun_loftr_2021}, backbone-agnostic keypoint/descriptor heads (SiLK)~\cite{gleize_silk_2023}, and temporal feature learning with IMUs (L-DYNO)~\cite{singh_l-dyno_2024}.

\subsection{Uncertainty in Visual Measurements}
Though disparate from feature learning, recent efforts to incorporate uncertainty into front-end systems have gained attention. Spatial uncertainty is modeled with multivariate Gaussian distributions of feature-space similarities~\cite{zaffar_estimation_nodate} or inverting factor graph covariances~\cite{belter_improving_2016, belter_modeling_2018}. Feature matching uncertainty using statistical filter pose prediction variance is measured in omnidirectional SLAM~\cite{valiente_robust_2017}. Semantically Informed Visual Odometry (SIVO)~\cite{ganti_network_2019} leverages MC-Dropout in a Bayesian UNet to select features based on segmentation class prediction variance. DistributionNet~\cite{yu_robust_2019} applies Gaussian distributions to person Re-Identification, while point clouds are randomly perturbed to measure spatial variance in~\cite{anderson_real-time_2019}. Pixel-wise uncertainty is estimated in neural radiance fields~\cite{sandstrom_uncle-slam_2023}, while ensembles of diffusion networks mimic predictive distributions to measure variance in~\cite{chan_hyper-diffusion_2024}.

Previous work has focused on incorporating uncertainty into back-end estimation using visual measurements but has overlooked uncertainty in the overall visual space, which could improve downstream estimation by providing a unified trust in visual data rather than focusing on individual feature matches, depth errors, or calibration issues. While hand-crafted feature-point algorithms lack this capability, recent learning-based methods provide a foundation for measuring general visual uncertainty. These methods implicitly learn to identify distinct, re-identifiable areas, offering a notion of \textit{interestingness} that could be extracted for such an estimate. Incorporating a sense of holistic uncertainty in the visual space before any subsequent visual measurements would benefit visual tasks by inherently providing a level of confidence in the visual information as a whole.
\section{\qq{}}\label{method}
\qq{} provides a temporal-free measure of visual information utility or ``featureness'' - the inherent interest and reliability of a pixel for robust feature extraction and description, instrumental to many autonomous systems tasks.  While humans intuitively understand which regions in an image are more visually significant, there is no universal definition of featureness that applies to all feature extractors and descriptors. This can lead to discrepancies in keypoint detection as demonstrated in~\autoref{fig:intro} (top left), where SIFT (blue) and ORB (green) identify distinct sets of keypoints within the same image. Current feature detection algorithms lack the understanding of stability and region-based visual consistency unless explicitly designed and trained for. We hypothesize that these properties are inherently captured during the training procedure of learning-based interest point detectors, and exploit this process through a multi-faceted approach. Additionally, we argue that uncertainty in the feature space is paramount to discerning visual utility, a formulation that the prior work has only alluded to. Bayesian solutions that provide this uncertainty through MC-Dropout inference are unsuitable for embedded systems where real-time or near-real-time operation is paramount, as multiple forward passes of networks are computationally expensive.

We describe \qq{} in this section, beginning with an overview of our baseline model and uncertainty head in~\autoref{sec:method-model}, followed by the training procedure in~\autoref{sec:method-train}, and finally our inference method to discern and integrate featureness in~\autoref{sec:method-inf}.

% Model architecture
\subsection{\qq{} Architecture}\label{sec:method-model}
Our baseline architecture is adapted from SiLK~\cite{gleize_silk_2023}, starting with a backbone-agnostic feature extractor and keypoint/descriptor heads. The keypoint head produces dense, pixel-wise probabilities representing how ``interesting'' a pixel is, following the SiLK definition, where a pixel is considered interesting if it can survive round-trip nearest-neighbor matching descriptions on transformed image pairs. Feature maps output from the backbone and the dense probability map output from the keypoint head are fed into a specialized uncertainty head, which outputs a normalized (0-1) per-pixel value representing a generalized conception of statistical Bayesian uncertainty (where 0 is certain and 1 is uncertain). The descriptor head is only used during training to refine the SiLK concept of interestingness (described more in~\autoref{sec:method-train}) and is not utilized during testing. This amounts to an inference \qq{} model that is incredibly lightweight, totaling one million parameters and 13 MB in model size.

\renewcommand{\thefigure}{2}
\begin{figure*}
    \centering
    \includegraphics[trim=1.25cm 1.25cm 1.25cm 1.25cm, clip=true,width=\linewidth]{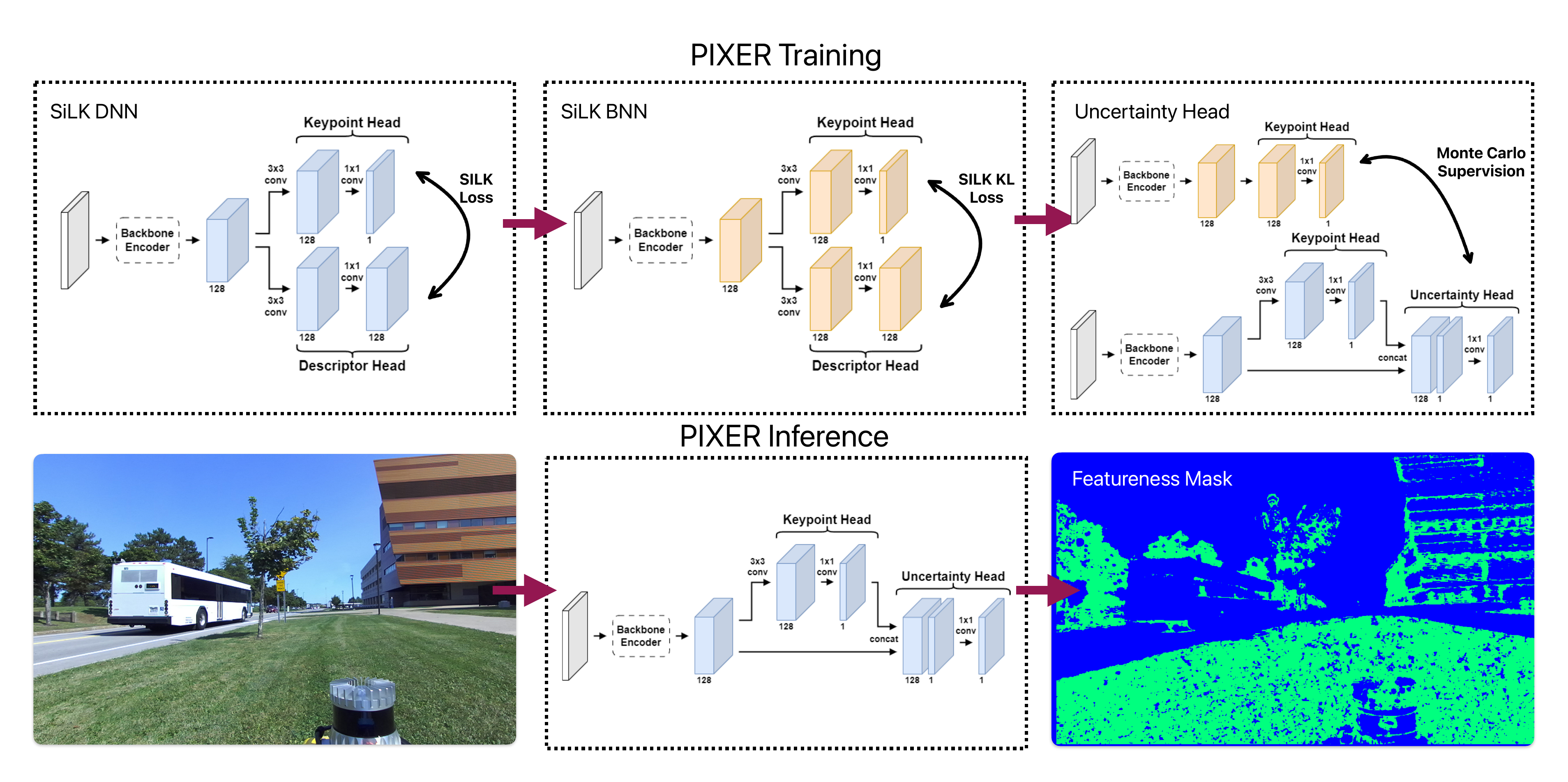}
    \caption{The training of \qq{} is a three-step process. First, we train a network with a general understanding of interestingness (i.e., feature point detection) where we make use of SiLK~\cite{gleize_silk_2023} in this work (top left). Next, we convert this model to a Bayesian Neural Network (BNN) and train again using the addition of probabilistic losses (e.g., KL Divergence~\cite{gal_dropout_2016}, top middle). Finally, we train a specialized \textit{uncertainty head} using feature variance computed by Monte Carlo supervision from the BNN (top right). The \qq{} inference model is then the joint feature-point probability and uncertainty networks (bottom middle). The combination of pixel-wise probability and uncertainty forms our definition of featureness $F$ (bottom right), used to describe the general utility of the visual information.
    % The \qq{} network is formed by connecting the uncertainty head with the deterministic SILK model. Top shows training mechanism while bottom shows inference, for each input image and thresholds $p_t$,$\sigma_t$ \qq{} outputs a featureness mask $F$}
    }
    \label{fig:network}
\end{figure*}

\subsection{\qq{} Training}\label{sec:method-train}
The training of \qq{} involves three distinct but correlated training rounds. First, the SiLK model is trained for feature point detection using the provided SiLK loss~\cite{gleize_silk_2023}, resulting in a general understanding of feature-point interestingness. The model is then converted into a Bayesian Neural Network (BNN) and undergoes an additional training round with a modified SiLK KL divergence loss: simply the addition of SiLK and KL Divergence~\cite{gal_dropout_2016} losses. This BNN outputs stochastic pixel-wise probability predictions where variance can be captured by multiple forward passes through the model, i.e. with Monte Carlo simulations. For the next stage, the probabilistic weights of the Bayesian SiLK model are frozen and the uncertainty head is trained in isolation. The uncertainty head takes as input the feature maps and dense probabilities output from a deterministic SiLK model (the first training round) and produces a pixel-wise uncertainty map. Using keypoint variances from Bayesian SiLK as a supervision signal (the second training round), the Binary Cross Entropy loss is employed to align the uncertainty. Afterward, a complete model is formed by connecting the trained uncertainty head with deterministic SiLK as shown in~\autoref{fig:network}. It is not strictly required to obtain these probabilities and feature maps from SiLK; uncertainty head learning can leverage any such feature technique that learns per-pixel probabilities of interestingness in a convolutional-based architecture, granted that the architecture supports Bayesian conversion~\cite{gal_dropout_2016}. In this work, we choose an in depth study of SiLK as the baseline provider to \qq{} due to its relative simplicity and overwhelming performance improvements to other such methods such as SuperPoint~\cite{detone_superpoint_2018}, R2D2~\cite{revaud_r2d2_2019}, and DISK~\cite{tyszkiewicz_disk_2020}.

% To mitigate this we introduce an ``uncertainty head'' which is trained to predict the variance of feature predictions directly from the BNN's output. This head learns from multiple Monte Carlo simulations and simplifies the process of calculating uncertainty, providing both probability and variance estimates for each pixel in an image. This output probability and variance together contribute to the image’s featureness.

% The SiLK model is initially trained for interestingness point detection using the standard SiLK loss, enabling it to learn a general understanding of interestingness. 

% The model is then converted into a Bayesian Neural Network (BNN) and undergoes further training with a modified SiLK KL divergence loss. 

% This allows the model, through Monte Carlo sampling, to provide per-pixel variance on interestingness. 

% Featureness filtering/reducing feature space
\subsection{Featureness Calculation}\label{sec:method-inf}
An input image is processed through our network to produce dense (i.e., pixel-wise) probability map $P$ and uncertainty map $U$, both scaled between 0 and 1. Our definition of featureness $F$ is calculated by the desired amount of probability ($p_t$) and uncertainty ($\sigma_t$) tolerance respectively, amounting in a binary selection mask:
\begin{equation}
F(x, y) =
\begin{cases}
1, & \text{if } P(x, y) \geq p_t \text{ and } U(x, y) \leq \sigma_t \\
0, & \text{otherwise}
\end{cases}
\label{eq:mask}
\end{equation}
for all $(x,y)$ pixel locations in the image. The settings of $p_t$ and $\sigma_t$ are designed to be application-specific, as we theorize that different applications that employ visual sensing may have different tolerance requirements. These parameters can thus be tuned empirically or conversely learned, although we primarily focus on the former in this work. The settings of $p_t$ and $\sigma_t$ can also be generalized through an expectation that high probability and low uncertainty pixels will be universally desired in any vision task. These properties are visually demonstrated in~\autoref{fig:intro} (bottom) with $P$, $U$, and $F$ heatmaps. Areas characterized by low texture and intensity variation typically exhibit low $P$ values, while regions with diverse textures, illumination, and intensity display high $P$. $U$, however, can hold different implications; a low $U$ doesn't necessarily equate to high featureness if accompanied by a low $P$. Conversely, a combination of low $U$ and high $P$ strongly suggests high featureness.

% A high $P$ and low $U$ signify that a pixel is more likely to be interesting and possesses high featureness. Conversely, low $P$ and high $U$ indicate low featureness.~\autoref{fig:intro} presents $P$ and $U$ heatmaps, visually depicting regions of high/low $P$ and $U$. 
% \red{This provides an intuition to how~\q

The featureness mask $F$ serves as a versatile estimate on the utility of visual information and can be integrated downstream in many forms. For example, $F$ can be used to selectively drop raw features detected in an image or by reducing the image space entirely. Both of these methods are front-end feature agnostic and operate on the visual information as a whole before any subsequent processing. Such a formulation permits a generalized design and conveys the methodology of feature-point learning without making use of any hard-chosen feature algorithm. This concept is instrumental to discerning visual utility, where $F$ can be summarized to take a confidence in the visual measurements.

% Featureness mask $F$ can be utilized in various forms, including for raw feature selection or by reducing the image space entirely. For feature selection, only features that lie in regions of the mask with high probability and low uncertainty, as determined by the thresholds, are retained. This method is agnostic to the feature detector and compatible with any front-end method. Alternatively, the image space can be reduced before feature extraction by filtering out raw image pixels using the mask, ensuring that no feature detections occur in uncertain or low-probability regions. 

% A trained~\qq{} model generates probability ($P$) and uncertainty ($U$) values on a per-pixel basis. These values are normalized between 0 and 1, representing the likelihood of a pixel being considered "interesting."  
% q{} reasons about the visual information utility of the given image.}
% \include{sections/tables/vo-eval}
% \include{sections/tables/in-out}
\section{Evaluation}\label{eval_section}
To demonstrate the effectiveness and versatility of \qq{} in autonomous systems and strive towards perception safety, we evaluate the performance on the visual odometry (VO) task across three environments with varying levels of dynamics and feature spaces including road scenes (KITTI~\cite{geiger_vision_2013}), urban campus traversals (CODa~\cite{zhang_toward_2024}), and a newly collected suburban campus dataset called \textit{Davis}. We start by describing the datasets in~\autoref{sec:datasets}, followed by our implementation for VO evaluation in~\autoref{sec:eval-imp}, and finally analyze the results in~\autoref{sec:eval-vo}.

\subsection{Datasets}\label{sec:datasets}
The KITTI dataset, consisting of urban and rural driving scenes, is used with randomly sampled images from sequences 11-21 for training and the near-static sequence 00 for testing. The CODa dataset features a small-wheeled robot navigating a highly dynamic urban campus environment. Here, we train on the medium dynamic sequence 22 and test on the high dynamic sequence 20, hypothesizing that our definition of featureness should inherently filter out a level of dynamic features that exhibit high uncertainty and, consequently, low featureness scores. Additionally, we capture a dataset using a ZED 2i stereo camera and ground-truth poses from a Mosaic X5 GNSS receiver mounted on a Boston Dynamics Spot robot. We collect sequences by walking the quadruped around a suburban campus environment, which exhibits a much sparser feature space than urban KITTI or CODa, in which high quality featureness should play a paramount role in VO estimation. The robot setup and a sample dataset image can be seen in~\autoref{fig:real-robot}. To ensure tracking stability and avoid problematic temporal moments (such as harsh rotations), we leverage the first 700 frames for our KITTI sequence, frames 1150 to 2200 for CODa, and frames 100 to 1500 for Davis.

% To assess the real-world applicability of our method, we capture images using a ZED 2i stereo camera and ground-truth poses from a Mosaic X5 GNSS reciever mounted on a Boston Dynamics Spot robot. \st{Ground truth pose estimations are obtained from ZED SDK which performs SLAM with stereo cameras providing much better estimation performance.} We collect sequences by walking the quadruped around a campus environment, an example image of the setup and sameple dataset image can be seen in~\autoref{fig:real-robot}  
% \task{Change GT method if GPS sequence doesn't work} 

\subsection{Implementation Details}\label{sec:eval-imp}
\renewcommand{\thefigure}{3}
\begin{figure}[htbp]
    \centering
    \includegraphics[width=\columnwidth]{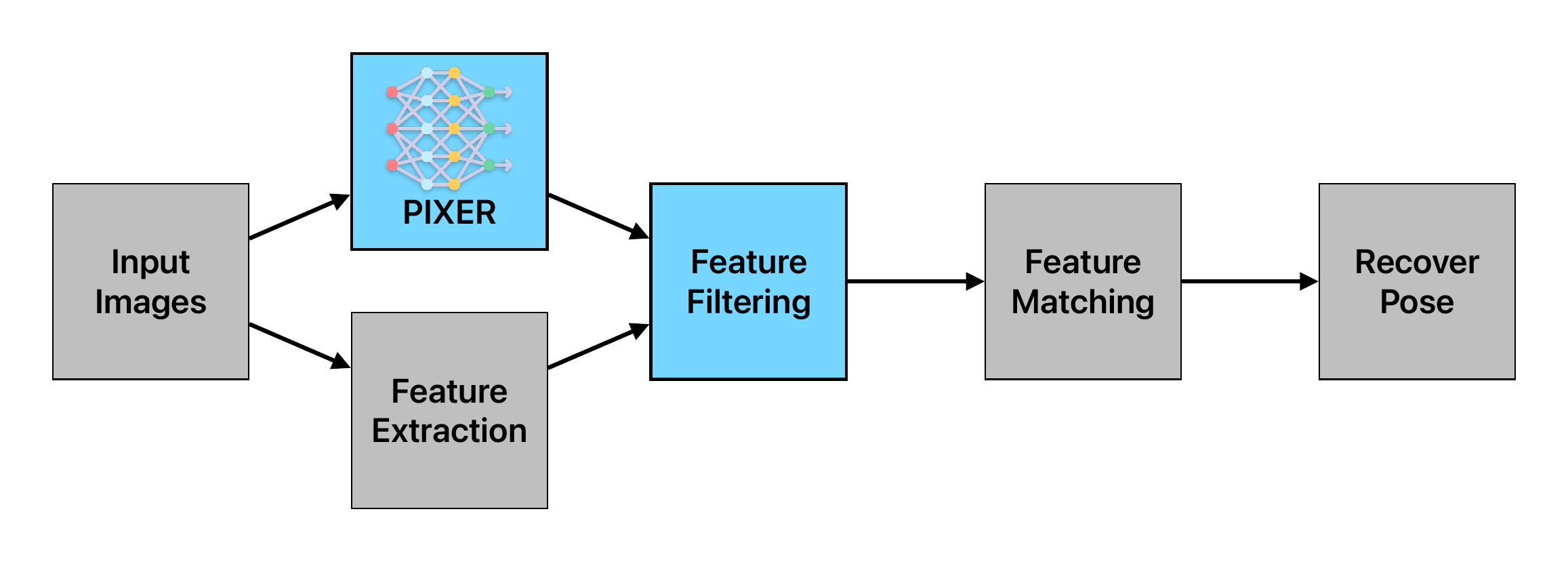}
    \caption{Visual Odometry pipeline (grey blocks) with \qq{} feature filtering based on featureness masks $F$ (blue blocks). 
    % $F$ is generated from input images which are then used to filter features.
    }
    \label{fig:pipeline}
\end{figure}

We train two sets of \qq{} models using KITTI and CODa imagery as described in~\autoref{sec:method-train}, and perform a generality experiment by applying the CODa model to the Davis dataset. Training is conducted on two NVIDIA A6000 GPUs with full-size images from KITTI (1241x376), CODa (1224x1024), and Davis (960x540), while inference is performed on an NVIDIA RTX 4090, achieving \textbf{average inference times of 1.93 ms}. To enhance pose estimation performance, we integrate~\qq{} into a monocular VO pipeline based on PySLAM~\cite{freda_luigifredapyslam_2024}, chosen for its efficiency in evaluating diverse features. We modify PySLAM to include feature filtering, as shown in~\autoref{fig:pipeline}, allowing the selective use of features from high-featureness regions. 

First, the featureness of the input image is estimated, generating a featureness mask $F$ as detailed in~\autoref{sec:method-inf}. We set $p_t=0.0$ and $\sigma_t=0.15$ for CODa and Davis, and $p_t=0.0$ and $\sigma_t=0.10$ for KITTI. $p_t$ is set to zero as we observed that the probability range output from SiLK is extremely variable (i.e., maximum probability values are not one). In many cases, this leads to settings of non-zero $p_t$ contributing to over-aggressive masking and feature removal as average pixel probabilities are incredibly low (e.g., 5\%), resulting in matching failures. This is a detriment of SiLK where we would normally expect every probability map to be uniformly scaled for each input image. Our featureness masks for VO evaluation are thus defined by removing only the \textit{most uncertain} pixels where $P > 0$.

% This is generally seen when using feature extractors that often output less features, we noticed that when using FAST and Shi-Tomasi, both of which are corner detectors, aggressive masking was usable but for the sake of consistency we choose to use a zero $p_t$.

Next, features extracted by PySLAM’s front-end are filtered based on this mask. We test the method on eight feature extractors, including hand-crafted (BRISK~\cite{leutenegger_brisk_2011}, FAST~\cite{rosten_fusing_2005}, ORB~\cite{rublee_orb_2011}, SIFT~\cite{lowe_object_1999}, Shi-Tomasi~\cite{jianbo_shi_good_1994}) and learned (SuperPoint~\cite{detone_superpoint_2018}, LOFTR~\cite{sun_loftr_2021}, SiLK~\cite{gleize_silk_2023}) features. The VO pipeline estimates the transformation between consecutive images through feature matching using filtered features selected by $F$ and the relative camera pose is recovered.

\subsection{Experimental Results}\label{sec:eval-vo}
\begin{table*}[]
\resizebox{\textwidth}{!}{
\begin{tabular}{@{}l|cccc|cccc|cccc@{}}
\toprule
\multirow{2}{*}{Feature} & \multicolumn{4}{c}{\textbf{KITTI} ($mean(F_\text{Area})$: 30.98\%)}& \multicolumn{4}{c}{\textbf{CODa} ($mean(F_\text{Area})$: 33.65\%)}& \multicolumn{4}{c}{\textbf{Davis} ($mean(F_\text{Area})$: 66.21\%)}\\ %\cmidrule(l){2-16}%
 & \multicolumn{1}{l}{RMSE (m)}  & \multicolumn{1}{l}{Time (ms)} & \multicolumn{1}{l}{$KP_\text{mean}$} & \multicolumn{1}{l|}{$KP_\text{mean}^\%$} & \multicolumn{1}{l}{RMSE (m)}  & \multicolumn{1}{l}{Time (ms)} & \multicolumn{1}{l}{$KP_\text{mean}$} & \multicolumn{1}{l|}{$KP_\text{mean}^\%$} & \multicolumn{1}{l}{RMSE (m)}  & \multicolumn{1}{l}{Time (ms)} & \multicolumn{1}{l}{$KP_\text{mean}$} & \multicolumn{1}{l}{$KP_\text{mean}^\%$}\\\midrule
 
BRISK & \phantom{0}1.78 & \textbf{\phantom{0}60} & \phantom{0,}752.34 & \phantom{0} & 
    46.64 & 116 & \phantom{00,}567.94 & \phantom{0} & 
    32.63 & \textbf{\phantom{0}91} & 1,125.75 & \phantom{0} \\
BRISK + Ours & \textbf{\phantom{0}1.55} & \textbf{\phantom{0}60} & \textbf{\phantom{0,}342.89} & 54.42 & 
    \textbf{46.33} & \textbf{114} & \textbf{\phantom{00,}180.81} & 68.20 & 
    \textbf{13.10} & \phantom{0}93 & \textbf{\phantom{0,}590.33} & 47.56\\\midrule

FAST & \phantom{0}2.63 & \phantom{0}53 & \phantom{0,}331.12 & \phantom{0} & 
    \phantom{0}9.37 & \phantom{0}68 & \phantom{00,}191.27 & \phantom{0} & 
    13.53 & 122 & \phantom{0,}869.57 & \phantom{0}\\
FAST + Ours & \textbf{\phantom{0}1.73} & \textbf{\phantom{0}50} & \textbf{\phantom{0,}185.23} & 44.06 & 
    \textbf{\phantom{0}4.39} & \textbf{\phantom{0}59} & \textbf{\phantom{00,0}97.63} & 48.96 & 
    \textbf{\phantom{0}8.84} & \textbf{118} & \textbf{\phantom{0,}602.64} & 30.70\\\midrule

LOFTR & 24.52 & 194 & 4,808.87 & \phantom{0} & 
    \textbf{10.14} & 451 & 16,661.33 & \phantom{0} & 
    \textbf{35.69} & 354 & 5,820.45 & \phantom{0}\\
LOFTR + Ours & \textbf{21.62} & \textbf{189} & \textbf{3,227.65} & 32.88 & 
    12.16 & \textbf{434} & \textbf{10,548.44} & 36.69 & 
    37.30 & \textbf{337} & \textbf{4,287.66} & 26.33 \\\midrule

ORB & \phantom{0}1.80 & \phantom{0}60 & \phantom{0,}512.54 & \phantom{0} & 
    47.10 & \phantom{0}83 & \phantom{00,}492.19 & \phantom{0} & 
    53.83 & \textbf{101} & \phantom{0,}654.07 & \phantom{0}\\
ORB + Ours & \textbf{\phantom{0}1.47} & \textbf{\phantom{0}59} & \textbf{\phantom{0,}281.84} & 45.01 & 
    \textbf{35.58} & \textbf{\phantom{0}77} & \textbf{\phantom{00,}154.23} & 68.66 & 
    \textbf{17.53} & 105 & \textbf{\phantom{0,}385.00} & 41.14\\\midrule

SIFT & \phantom{0}4.00 & \textbf{\phantom{0}62} & \phantom{0,}480.22 & \phantom{0} & 
    \phantom{0}6.57 & 185 & \phantom{0}1,205.78 & \phantom{0} & 
    29.20 & \textbf{\phantom{0}58} & \phantom{0,}300.08 & \phantom{0} \\
SIFT + Ours & \textbf{\phantom{0}1.82} & \phantom{0}63 & \textbf{\phantom{0,}206.48} & 57.00 & 
    \textbf{\phantom{0}2.69} & \textbf{179} & \textbf{\phantom{00,}411.52} & 65.87 & 
    \textbf{20.47} & \phantom{0}61 & \textbf{\phantom{0,}121.19} & 59.61\\\midrule

SuperPoint & \phantom{0}1.75 & \textbf{\phantom{0}65} & \phantom{0,}355.84 & \phantom{0} & 
    \phantom{0}4.29 & 133 & \phantom{0}1,708.62 & \phantom{0} & 
    33.83 & 150 & \phantom{0,}534.23 & \phantom{0}\\
SuperPoint + Ours & \textbf{\phantom{0}1.19} & \phantom{0}69 & \textbf{\phantom{0,}213.34} & 40.04 & 
    \textbf{\phantom{0}2.03} & \textbf{133} & \textbf{\phantom{00,}972.46} & 43.09 & 
    \textbf{27.63} & \textbf{148} & \textbf{\phantom{0,}355.84} & 33.39\\\midrule

% SURF & 2.82 & 2.22 & \textbf{76} & 462.66 & \phantom{0} & 56.13 & 44.96 & \textbf{237} & 346.96 & \phantom{0} & -1.00 & -1.00 & -1.000 & -1.00 & -1.00\\
% SURF + Ours & \textbf{1.22} & \textbf{1.07} & 77 & \textbf{207.16} & 55.22 & \textbf{38.49} & \textbf{23.81} & 238 & \textbf{\phantom{0}90.85} & 73.82 & -1.00 & -1.00 & -1.000 & -1.000 & -1.00\\\midrule

Shi-Tomasi & 30.16 & \textbf{\phantom{0}15} & \phantom{0,}385.94 & \phantom{0} & 
    43.18 & \textbf{\phantom{0}14} & \phantom{00,}497.58 & \phantom{0} & 
    28.95 & \textbf{\phantom{0}28} & 1,415.69 & \phantom{0}\\
Shi-Tomasi + Ours & \textbf{\phantom{0}3.66} & \textbf{\phantom{0}15} & \textbf{\phantom{0,}148.88} & 61.42 & 
    \textbf{39.97} & \textbf{\phantom{0}14} & \textbf{\phantom{00,}155.62} & 68.72 & 
    \textbf{22.36} & \phantom{0}31 & \textbf{\phantom{0,}898.84} & 36.51\\\midrule

SILK & 37.25 & \textbf{382} & \phantom{0,}441.82 & \phantom{0} & 
    55.74 & 590 & \phantom{00,}676.29 & \phantom{0} & 
    33.07 & \textbf{375} & \phantom{0,}674.15 & \phantom{0}\\
SILK + Ours & \textbf{20.40} & \textbf{382} & \textbf{\phantom{0,}217.30} & 50.82 & 
    \textbf{51.33} & \textbf{588} & \textbf{\phantom{00,}168.75} & 75.05 & 
    \textbf{31.90} & 376 & \textbf{\phantom{0,}292.55} & 56.60\\\bottomrule

\end{tabular}
}
\caption{Visual odometry (VO) performance results. Features when filtered using \qq{} contribute to a lower RMSE (31\% on average across all datasets) and frame-to-frame execution time for VO estimation (0.63\% despite the inclusion of model inference). This enables using lighter, faster features like Shi-Tomasi while achieving performance better than SIFT (e.g., KITTI \& Davis). We see a considerable reduction in the number of keypoints in all datasets by roughly 49\% ($KP_\text{mean}^\%$ shows \% reduction while $KP_\text{mean}$ shows the mean number of keypoints extracted per image). $mean(F_\text{Area})$ is the average percentage of pixels masked with $F$.}
\label{tab:vo-eval}
\vspace{-0.5cm}

\end{table*}

\renewcommand{\thefigure}{4}
\begin{figure}[htbp]
    \centering
    \includegraphics[trim=1.5cm 1.5cm 1.5cm 1.25cm, clip=true,width=\columnwidth]{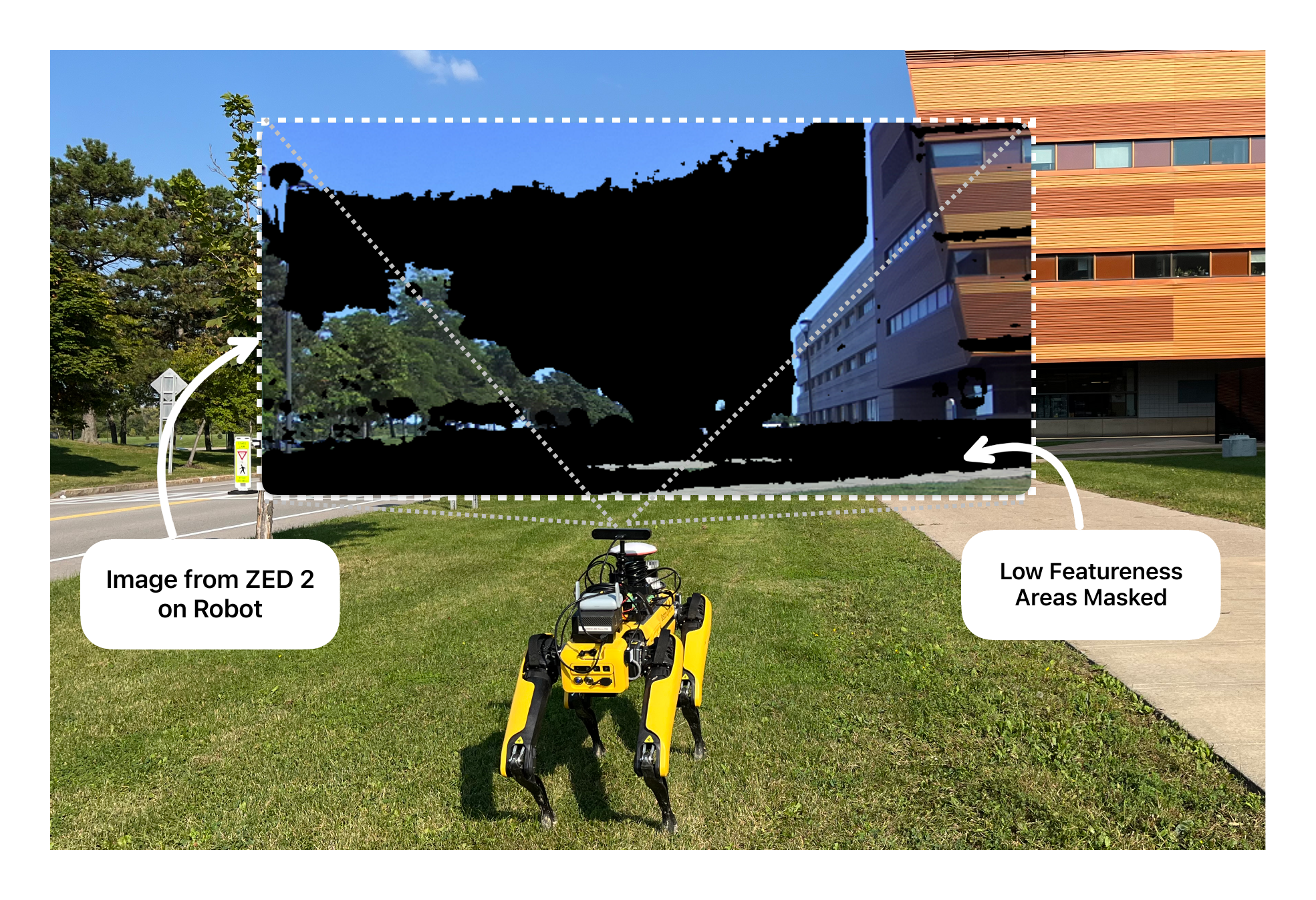}
    \caption{We evaluate \qq{} aided visual odometry on a custom dataset collected using a ZED 2i camera + Mosaic X5 GNSS on a Boston Dynamics Spot Quadruped. Results in 
    \autoref{tab:vo-eval} show superior estimation performance with mean RMSE improvement of 34\% and mean feature reduction of 41\%.}
    \label{fig:real-robot}
    \vspace{-0.5cm}
\end{figure}

% \subsubsection{Matching Performance}
% \autoref{tab:vo-eval} presents the mean and standard deviation of inliers and outliers identified during RANSAC feature matching. The proposed feature filtering technique leads to a significant reduction in the total number of features. Consequently, we observe a substantial decrease in outlier count during feature matching. This improvement can be contributed to the robustness and consistency of features filtered using~\qq{}.  The mean outlier count experiences a notable reduction of \red{XX\%}, while the standard deviation decreases by \red{XX}. While reducing the number of features inevitably leads to fewer inliers, this seemingly negative effect actually contributes to improved estimation performance and, more importantly, reduced compute time.

% \subsubsection{Estimation Performance on KITTI and CODa}
\autoref{tab:vo-eval} shows results of integrating our proposed \qq{} feature selection method on VO performance.  We measure trajectory accuracy using root mean square error (RMSE) comparing estimated trajectories with ground truth. Our experiments show that high-utility feature selection with \qq{} improves trajectory accuracy and processing speed across most feature types and datasets. This improvement stems from the robustness and stability of \qq{}'s selected features, which tend to correspond with regions of high featureness. By employing featureness masking (see~\autoref{sec:method-inf}), we effectively remove less informative features that hinder accurate matching and pose recovery. Feature reduction is measured in~\autoref{tab:vo-eval} where $KP_\text{mean}^s\%$ shows average reduction (\%) in number of feature keypoints. On the KITTI dataset, trajectory error decreases by an average of 37\%, with comparable or faster compute times and an average of 48\% fewer features used. Feature types with the highest RMSE are improved quite substantially, where we see roughly a 12\% and 88\% decrease on LOFTR and Shi-Tomasi respectively. This alludes to the importance of our featureness definition, which can limit over-reactive feature extractors and filter out low-quality detections. Additionally, we observe that SiLK is improved by roughly 45\%, indicating the importance of pixel-wise uncertainty information and the subsequent improvements despite using the same baseline architecture. Similar results are observed in the high-dynamics scene of CODa, with an average error improvement of 23\%, better processing time across all features, and an average reduction rate of 59\%. This demonstrates how the implicit featureness function captured by \qq{} can filter out dynamic pixels without any explicit training. Once again, we see substantial RMSE improvements on certain feature types including FAST (53\%), ORB (25\%), SIFT (59\%), and SuperPoint (53\%). Finally, for the Davis dataset, we see an average trajectory improvement of 34\% with an average of 41\% fewer features, revealing how high-utility feature selection is of paramount importance in feature-sparse scenes. \qq{} sees a slight increase in average frame time however (an average increase of 1.6\%), which could be due to greater feature extraction efficiency on smaller frame sizes where \qq{} still performs a network inference that is less affected by input shape. On this dataset, \qq{} improves trajectory estimation most substantially for BRISK (60\%), FAST (35\%), and ORB (67\%) features.

The VO results overall demonstrate \qq{}'s immediate utility by providing an average decrease of 31\% in trajectory error across all feature types and datasets, with an average improvement of 0.63\% in frame-to-frame computation time (including network inference) and 49\% less features used by the system.

\section{Discussion}\label{discussion}
{\bf Application}: The featureness definition estimated by \qq{} aims to mimic human intuition by identifying interesting regions within the feature space, where ``interesting'' is a generic and feature-agnostic concept. In an image, regions that contain textural and intensity variations are generally considered interesting, except when the texture is repetitive and has low visual utility - for example, a road surface, asphalt, glass and reflective surfaces, or any area exhibiting uniformity. \qq{} learns to measure this utility in a self-supervised manner, developing an intuition that exploits knowledge from feature-point learning. By combining learnable interest points with probabilistic Bayesian properties, \qq{} achieves a versatile notion of featureness applicable across various domains. The fusion of per-pixel probability and uncertainty allows for a customizable definition of visual utility, adjusting the balance between these factors as desired. This flexibility offers significant advantages for applications like VO, where even less effective feature detectors can be enhanced through high featureness selection, as demonstrated throughout the quantitative results in~\autoref{tab:vo-eval}. This supports broad applicability, as lightweight, less performative feature detectors can be boosted by \qq{}. \qq{} can additionally support downstream vision reasoning tasks such as semantic segmentation for highlighting import regions of the scene, where pixel-wise class predictions can be combined with featureness scores to reduce computation costs and improve perceptual capability. Furthermore, leveraging Bayesian reasoning in the visual utility measurement process leads to a reduction in information uncertainty, improving the agent's reliability and safety from the very initial entry of the perception system onward.

% that may not always align with human perception. 

% Through experiments, we observe parallels with what we, as trained perception researchers, might consider interesting, but there are cases where these parallels diminish. This implicit intuition is valuable, as it can enhance various vision applications such as structure-from-motion and 3D reconstruction, image stitching, visual place recognition, and medical imaging. Additionally, the featureness inferred by \qq{} can be propagated through the pose estimation pipeline, being used for efficient matching based on high featureness features or by incorporating the confidence of features into a probabilistic estimation back-end.

% \yash{Broadly applicably - use simpler/lighter feature detector (sift vs tomasi) or assist downstream tasks (SS) / safety reliability}
% \tj{Argue for using a poorer feature detector, if our results are truly correct, could use a worse detector and make it better via filtering.}
% \tj{now this is very broadly applicable, use a more lightweight feature detector and make it better, or use it downstream in apps like semantic segmentation that are meant to highlight more parts of the scene that are of interest. Assist with that with minimal cost of inference}
% \tj{need to mention safety, reason in a Bayesian way to minimize uncertainty and improve reliability and safety}

{\bf Limitations}: The main limitations of \qq{} are the probability ($p_t$) and uncertainty ($\sigma_t$) thresholds that define the featureness mask $F$. These parameters define the desired makeup of visual utility, which requires either empirical setting or automatic computation through learning or dataset statistics (which may be non-trivial). We expect general settings of featureness (e.g., high $p_t$ and low $\sigma_t$) to hold across datasets, tasks, and environments but can require re-tuning under edge cases or unique scenarios (such as the baseline feature learning mechanism outputting uniformly low probability estimates for all imagery). Additionally, the elongated training process of \qq{} is quite complex, which requires at minimum a convolutional feature network that is compatible with Bayesian learning. Although this work demonstrates the utility of \qq{} with SiLK,
%that can be used for any novel task or setup, 
one may desire a custom or more powerful expression of features that may not natively fit \qq{}'s training scheme or requirements.

% We identify some limitations of \qq{}, namely the probability $p_t$ and uncertainty $\sigma_t$ thresholds need to be empirically set which can be non-trivial. They generally work across datasets but can require re-tuning if the environment significantly changes. Also, \qq{}'s training is a multi-step process that needs a powerful compute setup, which can limit large-scale usability.

\section{Conclusion}
% \yash{dumping .......}

In this paper, we introduce \qq{}, a learning framework to estimate visual information utility or ``featureness''. We define featureness as the inherent interest and reliability of a pixel for robust feature extraction and description. \qq{} is completely self-supervised and uses a three-step training process including a Bayesian neural network which we generalize to provide single-shot uncertainty estimation avoiding expensive Monte Carlo simulations. We demonstrate \qq{}'s utility by incorporating feature filtering, i.e. picking features based on a featureness mask, into a visual-odometry pipeline. Our evaluation on three datasets and eight feature types show that visual utility estimation through \qq{} significantly reduces RMSE (avg. 31\%) while using fewer features (avg. 49\%). Overall, we believe \qq{} provides a novel, general measure of feature significance that has wide-ranging applications in robot perception.

\bibliographystyle{ieeetr} % specify a bibliography style that includes numbering 
\bibliography{root} % specify the name of your .bib file

\begin{thebibliography}{10}

\bibitem{barfoot_state_2017}
T.~D. Barfoot, {\em State {Estimation} for {Robotics}}.
\newblock Cambridge University Press, 1~ed., July 2017.

\bibitem{dellaert_factor_2017}
F.~Dellaert and M.~Kaess, ``Factor {Graphs} for {Robot} {Perception},'' {\em Foundations and Trends in Robotics}, vol.~6, no.~1-2, pp.~1--139, 2017.

\bibitem{detone_superpoint_2018}
D.~DeTone, T.~Malisiewicz, and A.~Rabinovich, ``{SuperPoint}: {Self}-{Supervised} {Interest} {Point} {Detection} and {Description},'' Apr. 2018.
\newblock arXiv:1712.07629 [cs].

\bibitem{christiansen_unsuperpoint_2019}
P.~H. Christiansen, M.~F. Kragh, Y.~Brodskiy, and H.~Karstoft, ``{UnsuperPoint}: {End}-to-end {Unsupervised} {Interest} {Point} {Detector} and {Descriptor},'' July 2019.
\newblock arXiv:1907.04011 [cs].

\bibitem{gleize_silk_2023}
P.~Gleize, W.~Wang, and M.~Feiszli, ``{SiLK} -- {Simple} {Learned} {Keypoints},'' Apr. 2023.
\newblock arXiv:2304.06194 [cs].

\bibitem{xiao_real-time_2024}
Z.~Xiao and S.~Li, ``A real-time, robust and versatile visual-{SLAM} framework based on deep learning networks,'' June 2024.
\newblock arXiv:2405.03413 [cs].

\bibitem{bloesch_robust_2015}
M.~Bloesch, S.~Omari, M.~Hutter, and R.~Siegwart, ``Robust visual inertial odometry using a direct {EKF}-based approach,'' in {\em 2015 {IEEE}/{RSJ} {International} {Conference} on {Intelligent} {Robots} and {Systems} ({IROS})}, pp.~298--304, Sept. 2015.

\bibitem{tanskanen_semi-direct_2015}
P.~Tanskanen, T.~Naegeli, M.~Pollefeys, and O.~Hilliges, ``Semi-direct {EKF}-based monocular visual-inertial odometry,'' in {\em 2015 {IEEE}/{RSJ} {International} {Conference} on {Intelligent} {Robots} and {Systems} ({IROS})}, pp.~6073--6078, Sept. 2015.

\bibitem{geneva_openvins_2020}
P.~Geneva, K.~Eckenhoff, W.~Lee, Y.~Yang, and G.~Huang, ``{OpenVINS}: {A} {Research} {Platform} for {Visual}-{Inertial} {Estimation},'' in {\em 2020 {IEEE} {International} {Conference} on {Robotics} and {Automation} ({ICRA})}, (Paris, France), pp.~4666--4672, IEEE, May 2020.

\bibitem{li_improving_2012}
M.~Li and A.~I. Mourikis, ``Improving the accuracy of {EKF}-based visual-inertial odometry,'' in {\em 2012 {IEEE} {International} {Conference} on {Robotics} and {Automation}}, pp.~828--835, May 2012.
\newblock ISSN: 1050-4729.

\bibitem{qin_vins-mono_2018}
T.~Qin, P.~Li, and S.~Shen, ``{VINS}-{Mono}: {A} {Robust} and {Versatile} {Monocular} {Visual}-{Inertial} {State} {Estimator},'' {\em IEEE Transactions on Robotics}, vol.~34, pp.~1004--1020, Aug. 2018.
\newblock Conference Name: IEEE Transactions on Robotics.

\bibitem{shan_lio-sam_2020}
T.~Shan, B.~Englot, D.~Meyers, W.~Wang, C.~Ratti, and D.~Rus, ``{LIO}-{SAM}: {Tightly}-coupled {Lidar} {Inertial} {Odometry} via {Smoothing} and {Mapping},'' in {\em 2020 {IEEE}/{RSJ} {International} {Conference} on {Intelligent} {Robots} and {Systems} ({IROS})}, pp.~5135--5142, Oct. 2020.
\newblock ISSN: 2153-0866.

\bibitem{shan_lego-loam_2018}
T.~Shan and B.~Englot, ``{LeGO}-{LOAM}: {Lightweight} and {Ground}-{Optimized} {Lidar} {Odometry} and {Mapping} on {Variable} {Terrain},'' in {\em 2018 {IEEE}/{RSJ} {International} {Conference} on {Intelligent} {Robots} and {Systems} ({IROS})}, (Madrid), pp.~4758--4765, IEEE, Oct. 2018.

\bibitem{campos_orb-slam3_2020}
C.~Campos, R.~Elvira, J.~J.~G. Rodr{\'\i}guez, J.~M.~M. Montiel, and J.~D. Tard{\'o}s, ``{ORB}-{SLAM3}: {An} {Accurate} {Open}-{Source} {Library} for {Visual}, {Visual}-{Inertial} and {Multi}-{Map} {SLAM},'' {\em arXiv:2007.11898 [cs]}, July 2020.
\newblock arXiv: 2007.11898.

\bibitem{cioffi_tightly-coupled_2020}
G.~Cioffi and D.~Scaramuzza, ``Tightly-coupled {Fusion} of {Global} {Positional} {Measurements} in {Optimization}-based {Visual}-{Inertial} {Odometry},'' in {\em 2020 {IEEE}/{RSJ} {International} {Conference} on {Intelligent} {Robots} and {Systems} ({IROS})}, (Las Vegas, NV, USA), pp.~5089--5095, IEEE, Oct. 2020.

\bibitem{dai_uncertainty-driven_2021}
X.-Y. Dai, Q.-H. Meng, and S.~Jin, ``Uncertainty-driven active view planning in feature-based monocular {vSLAM},'' {\em Applied Soft Computing}, vol.~108, p.~107459, Sept. 2021.

\bibitem{costante_uncertainty_2020}
G.~Costante and M.~Mancini, ``Uncertainty {Estimation} for {Data}-{Driven} {Visual} {Odometry},'' {\em IEEE Transactions on Robotics}, vol.~36, pp.~1738--1757, Dec. 2020.

\bibitem{gal_dropout_2016}
Y.~Gal and Z.~Ghahramani, ``Dropout as a {Bayesian} {Approximation}: {Representing} {Model} {Uncertainty} in {Deep} {Learning},'' Oct. 2016.
\newblock arXiv:1506.02142 [cs, stat].

\bibitem{ferianc_combinet_2021}
M.~Ferianc, D.~Manocha, H.~Fan, and M.~Rodrigues, ``{ComBiNet}: {Compact} {Convolutional} {Bayesian} {Neural} {Network} for {Image} {Segmentation},'' June 2021.
\newblock arXiv:2104.06957 [cs].

\bibitem{xu_cuahn-vio_2022}
Y.~Xu and G.~C. H.~E. de~Croon, ``{CUAHN}-{VIO}: {Content}-and-{Uncertainty}-{Aware} {Homography} {Network} for {Visual}-{Inertial} {Odometry},'' Aug. 2022.
\newblock arXiv:2208.13935 [cs].

\bibitem{stutts_lightweight_2023}
A.~C. Stutts, D.~Erricolo, T.~Tulabandhula, and A.~R. Trivedi, ``Lightweight, {Uncertainty}-{Aware} {Conformalized} {Visual} {Odometry},'' Mar. 2023.
\newblock arXiv:2303.02207 [cs, eess].

\bibitem{nir_designing_2024}
J.~S. Nir, D.~Giaya, and H.~Singh, ``On {Designing} {Consistent} {Covariance} {Recovery} from a {Deep} {Learning} {Visual} {Odometry} {Engine},'' Mar. 2024.
\newblock arXiv:2403.13170 [cs].

\bibitem{detone_toward_2017}
D.~DeTone, T.~Malisiewicz, and A.~Rabinovich, ``Toward {Geometric} {Deep} {SLAM},'' July 2017.
\newblock arXiv:1707.07410 [cs].

\bibitem{detone_self-improving_2018}
D.~DeTone, T.~Malisiewicz, and A.~Rabinovich, ``Self-{Improving} {Visual} {Odometry},'' Dec. 2018.
\newblock arXiv:1812.03245 [cs].

\bibitem{truong_glampoints_2019}
P.~Truong, S.~Apostolopoulos, A.~Mosinska, S.~Stucky, C.~Ciller, and S.~D. Zanet, ``{GLAMpoints}: {Greedily} {Learned} {Accurate} {Match} {Points},'' in {\em 2019 {IEEE}/{CVF} {International} {Conference} on {Computer} {Vision} ({ICCV})}, (Seoul, Korea (South)), pp.~10731--10740, IEEE, Oct. 2019.

\bibitem{revaud_r2d2_2019}
J.~Revaud, C.~De~Souza, M.~Humenberger, and P.~Weinzaepfel, ``{R2D2}: {Reliable} and {Repeatable} {Detector} and {Descriptor},'' in {\em Advances in {Neural} {Information} {Processing} {Systems}}, vol.~32, Curran Associates, Inc., 2019.

\bibitem{tyszkiewicz_disk_2020}
M.~Tyszkiewicz, P.~Fua, and E.~Trulls, ``{DISK}: {Learning} local features with policy gradient,'' in {\em Advances in {Neural} {Information} {Processing} {Systems}}, vol.~33, pp.~14254--14265, Curran Associates, Inc., 2020.

\bibitem{sun_loftr_2021}
J.~Sun, Z.~Shen, Y.~Wang, H.~Bao, and X.~Zhou, ``{LoFTR}: {Detector}-{Free} {Local} {Feature} {Matching} with {Transformers},'' Apr. 2021.
\newblock arXiv:2104.00680 [cs].

\bibitem{singh_l-dyno_2024}
K.~Singh, C.~Adhivarahan, and K.~Dantu, ``L-{DYNO}: {Framework} to {Learn} {Consistent} {Visual} {Features} {Using} {Robot}'s {Motion},'' in {\em 2024 {IEEE} {International} {Conference} on {Robotics} and {Automation} ({ICRA})}, pp.~17522--17528, May 2024.

\bibitem{zaffar_estimation_nodate}
M.~Zaffar, L.~Nan, and J.~F.~P. Kooij, ``On the {Estimation} of {Image}-matching {Uncertainty} in {Visual} {Place} {Recognition},''

\bibitem{belter_improving_2016}
D.~Belter, M.~Nowicki, and P.~Skrzypczynski, ``Improving accuracy of feature-based {RGB}-{D} {SLAM} by modeling spatial uncertainty of point features,'' in {\em 2016 {IEEE} {International} {Conference} on {Robotics} and {Automation} ({ICRA})}, (Stockholm), pp.~1279--1284, IEEE, May 2016.

\bibitem{belter_modeling_2018}
D.~Belter, M.~Nowicki, and P.~Skrzypczy{\'n}ski, ``Modeling spatial uncertainty of point features in feature-based {RGB}-{D} {SLAM},'' {\em Machine Vision and Applications}, vol.~29, pp.~827--844, July 2018.

\bibitem{valiente_robust_2017}
D.~Valiente, A.~Gil, L.~Pay{\'a}, J.~Sebasti{\'a}n, and {\'O}.~Reinoso, ``Robust {Visual} {Localization} with {Dynamic} {Uncertainty} {Management} in {Omnidirectional} {SLAM},'' {\em Applied Sciences}, vol.~7, p.~1294, Dec. 2017.

\bibitem{ganti_network_2019}
P.~Ganti and S.~L. Waslander, ``Network {Uncertainty} {Informed} {Semantic} {Feature} {Selection} for {Visual} {SLAM},'' in {\em 2019 16th {Conference} on {Computer} and {Robot} {Vision} ({CRV})}, pp.~121--128, May 2019.
\newblock arXiv:1811.11946 [cs].

\bibitem{yu_robust_2019}
T.~Yu, D.~Li, Y.~Yang, T.~Hospedales, and T.~Xiang, ``Robust {Person} {Re}-{Identification} by {Modelling} {Feature} {Uncertainty},'' in {\em 2019 {IEEE}/{CVF} {International} {Conference} on {Computer} {Vision} ({ICCV})}, (Seoul, Korea (South)), pp.~552--561, IEEE, Oct. 2019.

\bibitem{anderson_real-time_2019}
M.~L. Anderson, K.~M. Brink, and A.~R. Willis, ``Real-{Time} {Visual} {Odometry} {Covariance} {Estimation} for {Unmanned} {Air} {Vehicle} {Navigation},'' {\em Journal of Guidance, Control, and Dynamics}, vol.~42, pp.~1272--1288, June 2019.
\newblock Publisher: American Institute of Aeronautics and Astronautics.

\bibitem{sandstrom_uncle-slam_2023}
E.~Sandstr{\"o}m, K.~Ta, L.~Van~Gool, and M.~R. Oswald, ``{UncLe}-{SLAM}: {Uncertainty} {Learning} for {Dense} {Neural} {SLAM},'' in {\em 2023 {IEEE}/{CVF} {International} {Conference} on {Computer} {Vision} {Workshops} ({ICCVW})}, (Paris, France), pp.~4539--4550, IEEE, Oct. 2023.

\bibitem{chan_hyper-diffusion_2024}
M.~A. Chan, M.~J. Molina, and C.~A. Metzler, ``Hyper-{Diffusion}: {Estimating} {Epistemic} and {Aleatoric} {Uncertainty} with a {Single} {Model},'' Feb. 2024.
\newblock arXiv:2402.03478 [cs].

\bibitem{geiger_vision_2013}
A.~Geiger, P.~Lenz, C.~Stiller, and R.~Urtasun, ``Vision meets robotics: {The} {KITTI} dataset,'' {\em The International Journal of Robotics Research}, vol.~32, pp.~1231--1237, Sept. 2013.

\bibitem{zhang_toward_2024}
A.~Zhang, C.~Eranki, C.~Zhang, J.-H. Park, R.~Hong, P.~Kalyani, L.~Kalyanaraman, A.~Gamare, A.~Bagad, M.~Esteva, and J.~Biswas, ``Toward {Robust} {Robot} 3-{D} {Perception} in {Urban} {Environments}: {The} {UT} {Campus} {Object} {Dataset},'' {\em IEEE Transactions on Robotics}, vol.~40, pp.~3322--3340, 2024.
\newblock Conference Name: IEEE Transactions on Robotics.

\bibitem{freda_luigifredapyslam_2024}
L.~Freda, ``luigifreda/pyslam,'' Sept. 2024.
\newblock original-date: 2019-04-06T16:32:09Z.

\bibitem{leutenegger_brisk_2011}
S.~Leutenegger, M.~Chli, and R.~Y. Siegwart, ``{BRISK}: {Binary} {Robust} invariant scalable keypoints,'' in {\em 2011 {International} {Conference} on {Computer} {Vision}}, (Barcelona, Spain), pp.~2548--2555, IEEE, Nov. 2011.

\bibitem{rosten_fusing_2005}
E.~Rosten and T.~Drummond, ``Fusing points and lines for high performance tracking,'' in {\em Tenth {IEEE} {International} {Conference} on {Computer} {Vision} ({ICCV}'05) {Volume} 1}, (Beijing, China), pp.~1508--1515 Vol. 2, IEEE, 2005.

\bibitem{rublee_orb_2011}
E.~Rublee, V.~Rabaud, K.~Konolige, and G.~Bradski, ``{ORB}: {An} efficient alternative to {SIFT} or {SURF},'' in {\em 2011 {International} {Conference} on {Computer} {Vision}}, (Barcelona, Spain), pp.~2564--2571, IEEE, Nov. 2011.

\bibitem{lowe_object_1999}
D.~Lowe, ``Object recognition from local scale-invariant features,'' in {\em Proceedings of the {Seventh} {IEEE} {International} {Conference} on {Computer} {Vision}}, (Kerkyra, Greece), pp.~1150--1157 vol.2, IEEE, 1999.

\bibitem{jianbo_shi_good_1994}
{Jianbo Shi} and {Tomasi}, ``Good features to track,'' in {\em Proceedings of {IEEE} {Conference} on {Computer} {Vision} and {Pattern} {Recognition} {CVPR}-94}, (Seattle, WA, USA), pp.~593--600, IEEE Comput. Soc. Press, 1994.

\end{thebibliography}

\end{document}